\documentclass[preprint,12pt]{elsarticle}

\usepackage{amssymb}
\usepackage{amsmath}

\usepackage[utf8]{inputenc}
\usepackage{url}

\journal{Physics of Life Reviews}

\usepackage[firstpage]{draftwatermark}
\SetWatermarkText{Author's Accepted Manuscript (Postprint).\\
This manuscript has been accepted for publication in \textit{Physics of Life Reviews} (2025).\\
The final published version is available at:
\url{https://www.sciencedirect.com/science/article/pii/S1571064525001629} \\
Cite as: Hoffmann, M. (2025), 'The body is not there to compute: \\
Comment on “Informational embodiment: Computational role of information structure in codes and robots” by Pitti et al. ',\\ Physics of Life Reviews 55, 250-252.
}
\SetWatermarkScale{0.3}       
\SetWatermarkAngle{0}        
\SetWatermarkColor[gray]{0.5} 
\SetWatermarkVerCenter{0.08\paperheight}  
\SetWatermarkHorCenter{0.5\paperwidth}    

\begin{document}

\begin{frontmatter}



\title{The body is not there to compute \\
Comment on ``Informational embodiment: Computational role of information structure in codes and robots'' by Pitti et al.
}


\author{Matej Hoffmann} 

\affiliation{organization={Department of Cybernetics, Faculty of Electrical Engineering, \\ Czech Technical University in Prague}
            }








\end{frontmatter}



Applying the lens of computation and information has been instrumental in driving the technological progress of our civilization as well as in empowering our understanding of the world around us. The digital computer was and for many still is the leading metaphor for how our mind operates. Information theory (IT) has also been important in our understanding of how nervous systems encode and process information. The target article deploys information and computation to bodies: to understand why they have evolved in particular ways (animal bodies) and to design optimal bodies (robots). In this commentary, I argue that the main role of bodies is not to compute. 

\section{Morphological non-computation}
Embodiment or body morphology (the shape of the body and limbs, the type and placement of sensors and effectors) and the material properties of the elements composing the morphology have a key role in perception, behavior, and cognition \cite{PfeiferBongard2007}. The concept of ``morphological computation'' largely overlaps with embodiment, but, as we argue in detail in \cite{Mueller2017}, it introduces the word computation, together with perhaps undesired implications. In \cite{PfeiferBongard2007}, morphological computation has been introduced in a metaphorical sense: if the body has a perfect design for a task, like the passive dynamic walker (PDW) \cite{McGeer1990} for walking, it ``offloads'' computation from the brain or controller to the body. However, this is only a metaphor. The PDW is ``pure physics walking''. It has not been designed to compute and hence it does not compute \cite{Horsman2014}---unless everything computes (pancomputationalism).

Morphological computation has been taken beyond the offloading metaphor by some.  Paul \cite{Paul2006} designed a robot that ``computes'' the XOR function.  Hauser, Nakajima and colleagues showed that physical systems with rich body dynamics like masses and springs or soft bodies resembling octopus arms can act as physical reservoir computers emulating or computing the evolution of specific nonlinear dynamical systems (e.g., \cite{nakajima2015information}). However, such examples remain demonstrators rather than practical technological solutions. After all, the primary role of the body is to physically interact with the environment and not to compute; animals have evolved neural systems to compute. A properly designed body has a key role in generating behavior and simplifying control (``morphology facilitating control'' \cite{Mueller2017}), but its role is not a computational one. 

Rolf Pfeifer once told me that after a talk about morphological computation he had given, he was approached by Inman Harvey, who said: ``I agree with everything you said. Except, I would call it morphological non-computation.''

\section{The bliss of motor redundancy}
Pitti et al. in the target article argue that IT can be applied to perception, control, and body design. In my view, entropy maximization and efficient coding may prove the most useful for perception. In animals, the placement and properties of receptors have evolved to preprocess sensory information. The eye morphologies of different insects have evolved to aid different tasks such as keeping a distance from obstacles, escape or attack maneuvers. Maximizing entropy and hence minimizing redundancy in sensory streams seems useful. 

The target article attempts to apply the same principle to motor control and discusses the accuracy of control for different bodies and with different sensory and motor accuracy. Applying the principle of entropy maximization (PEM) to motor control would simultaneously minimize the redundancy. However, redundancy (or motor abundance) is actually a bliss \cite{latash2012bliss} and is responsible for the versatility and robustness of animals (which is probably a more important cost function than accuracy). The target article rightly mentions motor synergies as a way of coping with the redundancy. Yet, applying PEM to motor systems would likely reduce motor abundance too much and make organisms brittle when faced with unexpected situations. 

\section{Dynamical systems or information theory?}
The natural description of mechanical systems interacting with the world seem to be dynamical systems, also used by control theory. The dynamical systems description can deal with continuous time and space, nonlinear relations, phase changes etc. The description may not be tractable and hence useful when the dimensionality increases. IT methods like mutual information or Granger causality have been applied to high-dimensional streams like recordings from the brain. 

The target article applies information-theoretic measures like entropy to understand the relation of different robot body configurations and accuracy in tasks like reaching and grasping. Yet, such treatment inherently collapses the closed-loop interaction of the brain-body-environment in continuous space and time, with multiple nested time scales, into a simplified discretized representation. Moreover, even if such a mapping preserved the important characteristics of the interaction, IT provides at best indirect cost functions. 
To find the best robot controller, it seems more straightforward to use cost functions directly measuring task performance and employ variants of reinforcement learning. If set up properly, the robot will have the opportunity to discover solutions exploiting its embodied interaction with the environment. This can be combined with simulation, leveraging current progress in robot simulators and sim-to-real transfer techniques \cite{ju2022transferring}.

McGeer \cite{McGeer1990} has successfully employed dynamical systems to explain the operation and stability of the PDW. Designing optimal robot bodies can draw on modeling and control \cite{pekarek2010variational,zardini2021co} or deep evolutionary learning \cite{gupta2021embodied}. Perhaps the PDW can be a testbed for the framework proposed by the target article. Can IT be used to describe the operation or even design a PDW?

\section{Conclusion}
The target article builds on works on embodied intelligence and ``morphological computation'' and emphasizes the role of bodies in perception, control, and body design. This is especially timely now in the time of ``Embodied AI'' and foundation models for robotics---the idea of leveraging the success of transformer networks in the language and visual domains and extending them to robotics through so-called Vision-Language-Action models (VLAs) \cite{kawaharazuka2025vision}. I argue that the treatment of the robot embodiment in these architectures is very shallow (see \cite{hoffmann2026embodied} for details). 
In the era of AI and large neural network models, the target article brings embodiment back in the focus. 
However, I do not share the view of Pitti et al. that information theory (IT) will be instrumental in designing robot bodies or controllers.  Evolution may have optimized neural systems for efficient coding, but animal bodies have evolved for survival: to robustly act in the world with minimal energy expenditure (and not to compute). 

\section*{Acknowledgments}
This work was supported by the Czech Science Foundation (GA ČR), project no. 25-18113S.



\bibliographystyle{elsarticle-num} 
\bibliography{Hoffmann_commentaryOnPitti}





\end{document}